\documentclass{article} 
\usepackage{iclr2027_conference,times}
\usepackage{booktabs}
\usepackage{graphicx}
\usepackage{algorithm}
\usepackage{algpseudocode}
\usepackage{subcaption}
\usepackage{wrapfig}
\usepackage{booktabs}
\usepackage[table]{xcolor}

\definecolor{groupgray}{RGB}{242,242,242}
\definecolor{retttgreen}{RGB}{226,241,236}
\definecolor{retttorange}{RGB}{252,235,218}

\usepackage{amsmath,amsfonts,bm}

\def\eqref#1{equation~\ref{#1}}

\def\1{\bm{1}}

\DeclareMathAlphabet{\mathsfit}{\encodingdefault}{\sfdefault}{m}{sl}
\SetMathAlphabet{\mathsfit}{bold}{\encodingdefault}{\sfdefault}{bx}{n}

\usepackage{hyperref}
\usepackage{url}

\title{ReTTT: Reflection-Driven Test-Time Training via Experience-Based Parametric Memory}

\author{
\textbf{Kangcong Li}$^{1,2}$\thanks{Work done during an internship at Shanghai AI Lab.}\quad
\textbf{Hangfan Zhang}$^{2}$\thanks{Corresponding authors: \texttt{zhanghangfan@pjlab.org.cn}, \texttt{eetchen@fudan.edu.cn}.}\quad
\textbf{Shuyue Hu}$^{2}$\quad
\textbf{Shengji Tang}$^{3}$\quad
\textbf{Tao Chen}$^{1}$\footnotemark[2] \\[6pt]
$^{1}$Fudan University\quad
$^{2}$Shanghai Artificial Intelligence Laboratory\quad
$^{3}$CUHK
}

\iclrfinalcopy 
\begin{document}

\maketitle

\begin{abstract}
LLM agents accumulate valuable task-specific information through their actions, observations, and environmental feedback during multi-turn task execution.
When effectively used, this information can help agents refine subsequent decisions and improve task performance.
However, the learning value of an interaction depends on its context and subsequent feedback, making it challenging to extract reliable, actionable lessons and revise them as the task unfolds.
In this paper, we introduce ReTTT, which enables experience-driven self-adaptation through test-time training on lessons constructed and continually revised within an episode.
ReTTT lets the agent reflect on interaction feedback, maintain the resulting lessons in an explicit memory, and learn from them by updating an episode-local LoRA adapter while keeping the backbone frozen.
The adapted model then performs both subsequent actions and reflection, coupling policy adaptation with the construction of future learning experience.
We evaluate ReTTT on ALFWorld, ScienceWorld, and SWE-bench Lite using two model families.
ReTTT outperforms trajectory-based test-time training across all six model--benchmark settings, improving task success by up to 12.14 percentage points.
A frozen-reflection comparison further shows a 7.14-point benefit from adapting the reflection policy, while trajectory analysis illustrates how adapted reflection derives more task-specific guidance from failed attempts.
These findings suggest that learning to reflect on its own behavior can boost agents' performance on long-horizon tasks.
\end{abstract}

\section{Introduction}
\label{sec:introduction}

Modern large language model (LLM) agents~\citep{liu2024agentbench} tackle multi-turn tasks, such as software development~\citep{yang2024swe,jimenez2024swe}, web navigation~\citep{zhou2024webarena}, and scientific research~\citep{wang2022scienceworld}, through an iterative agent loop.
The agent alternates between reasoning, taking actions, and receiving environmental feedback, accumulating a trajectory of plans, intermediate reasoning, actions, and observations in its context~\citep{yao2022react}.
This trajectory records action outcomes and task progress, providing task-specific experience that can inform subsequent decisions and learning.

A substantial body of work investigates how agents can make better use of this interaction history.
Context management methods organize, compress, and retrieve historical information to guide subsequent actions through the input context while keeping model parameters fixed~\citep{park2023generative,packer2023memgpt,zhang2026adamem,mi2026procmem}.
Parametric memory approaches consolidate accumulated experience into learned parameters, allowing information from past trajectories to shape the agent's policy beyond the immediate context~\citep{yao2026parammem,ren2026scaling}.
Recently, agentic test-time training (aTTT) has brought parameter adaptation into ongoing task execution, using signals from agent outputs, environmental observations, or generated progress summaries~\citep{wang2026no}.
It downweights repeated token spans to limit redundant self-training while retaining novel information.
Each update changes the policy that acts next and thereby shapes the interaction history available for subsequent learning.

\begin{figure}[t]
    \centering
    \includegraphics[width=\linewidth]{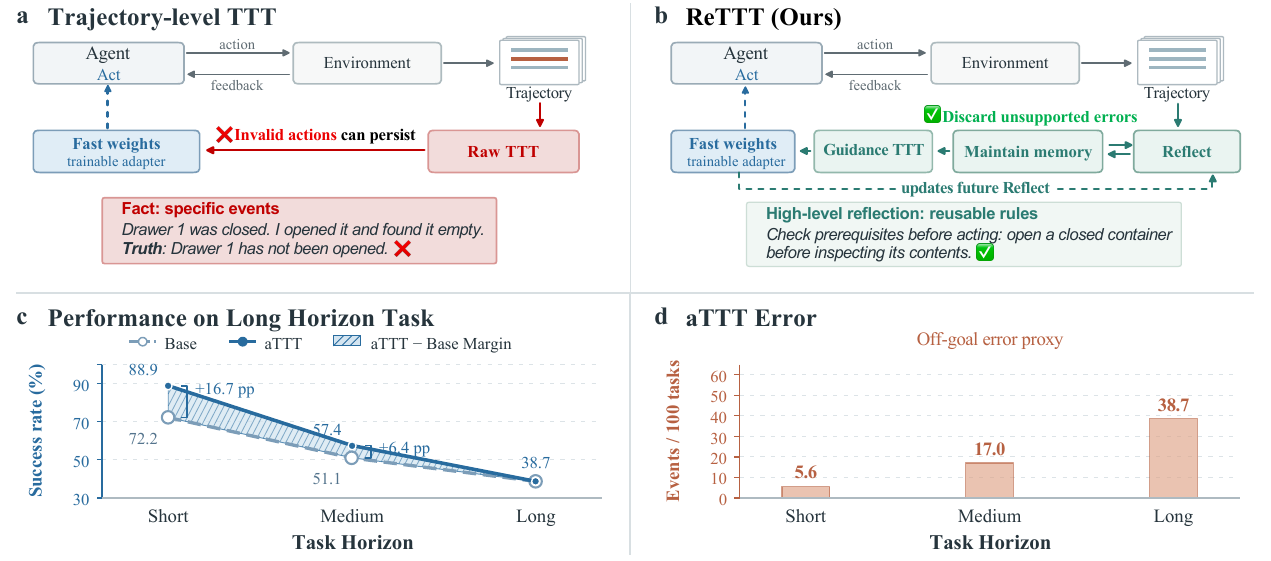}
    \caption{
\textbf{Motivation: interaction trajectories do not directly provide reliable supervision.}
\textbf{(a)} Trajectory-level TTT internalizes raw experience, allowing unsupported errors to persist.
\textbf{(b)} ReTTT filters procedural guidance before internalization and adapts reflection through shared fast weights.
\textbf{(c)} aTTT's gain over Base decreases from +16.7 pp to +6.4 pp and 0 with longer horizons.
\textbf{(d)} aTTT's off-goal error proxy increases from 5.6 to 17.0 and 38.7.
}
    \label{fig:motivation}
\end{figure}

However, continual parameter updates leave a key question: what should an agent learn from its ongoing interactions?
An interaction trajectory records what happened, but useful supervision must capture what should guide future behavior.
Failure feedback, for example, can turn an unsuccessful action into a lesson about prerequisites for subsequent attempts(Figure~\ref{fig:motivation}a).
New observations can also revise earlier conclusions, requiring the supervision used for later updates to change as the task unfolds.
The challenge is therefore to continually derive reliable, decision-relevant lessons from the trajectory and use them to adapt the policy within the same episode.

To investigate the limitations of agentic test-time training over longer task horizons, we analyze aTTT's performance and execution errors.
Its improvement over the frozen policy falls from 16.7 percentage points on short tasks to 6.4 on medium tasks and disappears on long tasks (Figure~\ref{fig:motivation}c).
Meanwhile, aTTT's off-goal error proxy, measured as placements of target objects at non-goal locations per 100 tasks, rises from 5.6 to 17.0 and 38.7 across these groups (Figure~\ref{fig:motivation}d).
These findings suggest that accumulating more interaction data does not necessarily provide better learning supervision, motivating the explicit construction and revision of task-relevant lessons before parameter updates.

To address these challenges, we introduce ReTTT(Figure~\ref{fig:motivation}b), which couples the construction and revision of learning experience with policy adaptation through test-time training.
The same adapted model acts in the environment and constructs the lessons used for later updates within the ongoing episode.
At periodic reflection steps, it interprets interaction feedback to extract task-relevant lessons, including action prerequisites, failure conditions, and guidance for remaining subgoals.
These lessons are checked against interaction evidence and maintained in an explicit memory that is revised as new information arrives.
New and revised lessons, together with selected prior experience, are then used to update an episode-local LoRA adapter through test-time training, while the backbone remains frozen.
We instantiate these lessons as question--answer pairs (ReTTT-QA) and structured summaries (ReTTT-Sum).

We evaluate ReTTT on ALFWorld, ScienceWorld, and SWE-bench Lite using Qwen3.5-9B and Gemma4-12B.
Both ReTTT-QA and ReTTT-Sum outperform aTTT in all six model--benchmark settings (Table~\ref{tab:main_results}).
With Qwen3.5-9B, ReTTT-QA improves success over aTTT by 12.14, 11.67, and 3.33 percentage points on the three benchmarks, respectively.
On ALFWorld, ReTTT-QA's advantage over aTTT grows from 5.6 percentage points on short tasks to 14.7 on long tasks, grouped by reference action count (Table~\ref{tab:alfworld_horizon}).

Further analysis examines how adapting the reflection policy contributes to these improvements.
On ALFWorld with Qwen3.5-9B, ReTTT-QA achieves 63.57\% success, compared with 56.43\% when the reflection policy remains frozen while action-policy adaptation is retained (Section~\ref{sec:closed_loop_exp}).
A trajectory-level example illustrates the difference: with the same interaction history, the adapted reflector derives task-specific search guidance from failed attempts, while the frozen reflector repeats generic advice (Figure~\ref{fig:reflection_trajectory}).
These observations suggest that adapting reflection helps the agent derive more actionable lessons from its experience, supporting continued progress during long-horizon task execution.


Our contributions are threefold:
\begin{itemize}
    \item We identify a mismatch between interaction history and useful learning supervision in agentic test-time training. Quantitative comparisons examine the learning signals used during adaptation and their relationship to reliable long-horizon task execution.
    \item We introduce ReTTT, which enables experience-driven self-adaptation through test-time training on lessons distilled and continually revised during task execution, with the adapted model responsible for both actions and subsequent reflection.
    \item Experiments across three benchmarks and two model families demonstrate higher task success than aTTT for both ReTTT variants. Frozen-reflection comparisons and trajectory analysis further provide evidence for the benefit of adapting the policy that constructs learning experience.
\end{itemize}

\section{Method}
\label{sec:method}

\begin{figure}[t]
    \centering
    \includegraphics[width=\linewidth]{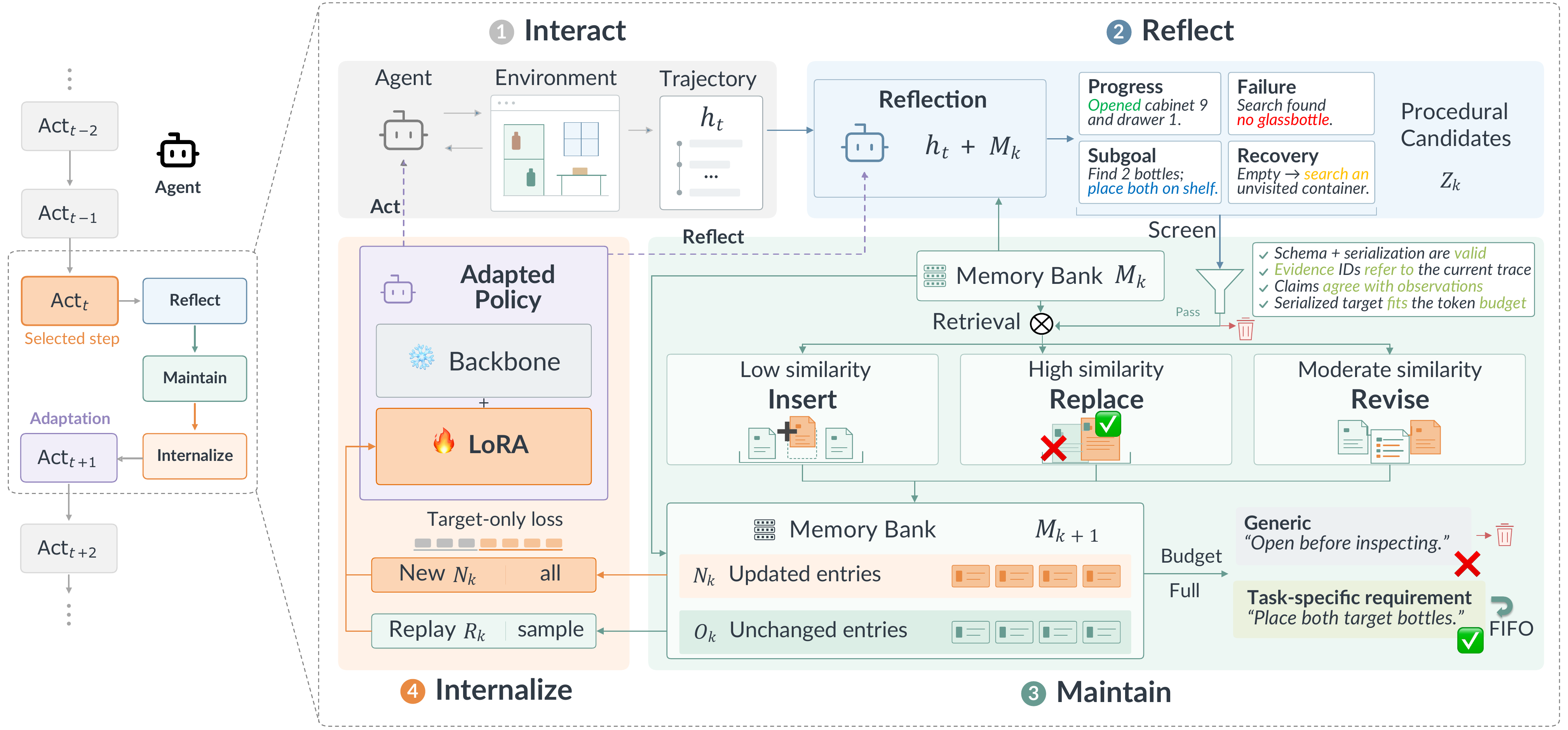}
    \caption{
\textbf{Overview of ReTTT.}
\textbf{(1) Interact:} the adapted policy acts in the environment.
\textbf{(2) Reflect:} interaction history is converted into procedural candidates $\mathcal{Z}_k$.
\textbf{(3) Maintain:} candidates are screened and reconciled with memory, producing changed entries $\mathcal{N}_k$ and unchanged entries $\mathcal{O}_k$.
\textbf{(4) Internalize:} $\mathcal{N}_k$ and replay $\mathcal{R}_k\subseteq\mathcal{O}_k$ update episode-local fast weights while the backbone remains frozen.
}
    \label{fig:method}
\end{figure}

Figure~\ref{fig:method} illustrates the data flow of ReTTT, and Algorithm~\ref{alg:rettt} summarizes the complete in-episode procedure.
The agent interacts with the environment using episode-local fast weights and periodically performs three adaptation stages: it reflects on the observed trajectory to construct procedural candidates, maintains these candidates in a bounded procedural memory, and internalizes newly acquired procedures together with replayed prior experience.
The backbone remains frozen throughout the episode, while action and reflection use two prompting interfaces to the same adapted model.

\subsection{Problem Formulation}
\label{sec:formulation}

We consider a long-horizon task as a partially observed interactive process.
At environment step $t$, the agent observes $o_t$, selects an action $a_t$, and receives the next observation $o_{t+1}$ together with feedback $r_t$.
The interaction history before selecting $a_t$ is
\begin{equation}
    h_t
    =
    (o_1,a_1,r_1,\ldots,a_{t-1},r_{t-1},o_t).
\end{equation}

We use $k$ to index adaptation events and let $t_k$ denote the environment step at which event $k$ occurs.
The backbone parameters $\theta$ remain frozen throughout an episode.
ReTTT maintains episode-local fast weights $\phi_k$, implemented as a lightweight trainable adapter, together with an explicit procedural memory $\mathcal{M}_k$.
Between adaptation events, actions are generated by
\begin{equation}
    a_t
    \sim
    \pi_{\theta,\phi_k}
    (\cdot \mid g,h_t),
    \label{eq:action_policy}
\end{equation}
where $g$ denotes the task specification and $\phi_k$ is the current fast-weight state.

Action generation and reflection are two prompting interfaces to the same adapted language model and therefore share $\theta$ and $\phi_k$.
The explicit memory $\mathcal{M}_k$ is supplied only to the reflection interface; action generation conditions on the interaction history and the procedural knowledge already internalized in $\phi_k$.
Both $\mathcal{M}_k$ and $\phi_k$ are reset at the end of each episode.

\subsection{Adapted-Policy Reflection}
\label{sec:reflection}

At adaptation event $k$, ReTTT constructs a structured reflection context from the task specification, the observed interaction history, and the current procedural memory:
\begin{equation}
    c_k^{\mathrm{ref}}
    =
    \mathrm{Reflect}
    \left(
        g,
        h_{t_k},
        \mathcal{M}_k;
        \mathcal{S}_{\mathrm{ref}}
    \right),
    \label{eq:reflection_prompt}
\end{equation}
where $\mathcal{S}_{\mathrm{ref}}$ denotes the reflection prompt.
The current adapted model then generates a set of candidate procedural experiences:
\begin{equation}
    \mathcal{Z}_k
    =
    \{z_{k,i}\}_{i=1}^{m_k}
    \sim
    R_{\theta,\phi_k}
    \left(
        \cdot
        \mid
        c_k^{\mathrm{ref}}
    \right).
    \label{eq:reflection_generation}
\end{equation}

The reflection organizes interaction experience into four procedural categories.
\textbf{Progress} records completed state transitions or subgoals supported by environment feedback.
\textbf{Failure} records unsuccessful or inadmissible transitions and their observed conditions.
\textbf{Remaining subgoals} captures unresolved requirements and missing prerequisites.
\textbf{Recovery} specifies reusable actions or recovery strategies.
Each item is accompanied by supporting trajectory evidence.
Accordingly, a candidate is represented as
\begin{equation}
    z_{k,i}
    =
    \left(
        \kappa_{k,i},
        u_{k,i},
        E_{k,i}
    \right),
    \qquad
    \kappa_{k,i}
    \in
    \{
        \textsc{Progress},
        \textsc{Failure},
        \textsc{Subgoal},
        \textsc{Recovery}
    \},
    \label{eq:candidate_procedure}
\end{equation}
where $u_{k,i}$ is the procedural content and $E_{k,i}$ contains its supporting trajectory evidence.
At this stage, $\mathcal{Z}_k$ contains only candidate experience; admission to memory is determined by the maintenance stage.

We instantiate the same procedural schema using two textual serializations.
In the QA representation, each item is written as a decision-relevant question, answer, and supporting evidence:
\begin{equation}
    z_{k,i}^{\mathrm{QA}}
    =
    \left(
        q_{k,i},
        a_{k,i}^{\mathrm{ans}},
        E_{k,i}
    \right).
    \label{eq:qa_reflection}
\end{equation}
In the Summary representation, the same information is organized as a structured checkpoint containing progress, failures, remaining subgoals, procedures, and supporting evidence.

\subsection{Procedural Memory Maintenance}
\label{sec:consolidation}

The candidate set $\mathcal{Z}_k$ is processed before it can be used for parameter updates.
ReTTT first screens the candidates for structural validity and consistency with the observed trajectory, discarding unsupported or malformed experience before memory integration.
The surviving candidates are then reconciled with the current memory $\mathcal{M}_k$:
\begin{equation}
\mathcal{M}_{k+1}
=
\mathrm{Maintain}
\left(
\mathcal{M}_k,\,
\mathrm{Screen}(\mathcal{Z}_k; h_{t_k})
\right).
\label{eq:memory_update}
\end{equation}

\textbf{Candidate screening.}
A candidate is retained only if it follows the expected procedural schema, its supporting evidence refers to observations in the current trajectory, and its content is consistent with the observed interaction and task outcome.
Candidates that fail these checks are discarded before they can enter memory or contribute to parameter updates.
We denote the surviving set by
\begin{equation}
\widetilde{\mathcal{Z}}_k
=
\left\{
z \in \mathcal{Z}_k
\;\middle|\;
\mathrm{Valid}(z, h_{t_k}) = 1
\right\},
\label{eq:screened_candidates}
\end{equation}
where $\mathrm{Valid}$ applies deterministic schema, evidence-grounding,
trajectory-consistency, and serialization checks.
Candidates whose serialized targets exceed the allowed memory-token budget
are also discarded.

\textbf{Memory reconciliation.}
For each surviving candidate, ReTTT computes its similarity to existing memory
entries based on normalized procedural type and decision-relevant content.
If the highest similarity is low, the candidate is inserted as a new entry.
If it is high, the most similar entry is replaced by the newly observed version.
For moderate similarity, the existing entry is revised by merging the old and
new information.
Entries unaffected by the current candidates remain unchanged.

After reconciliation and memory-budget enforcement, we denote by $\mathcal{N}_k$ the entries in $\mathcal{M}_{k+1}$ that were inserted, replaced, or revised at event $k$.
The remaining unchanged entries are
\begin{equation}
    \mathcal{O}_k
    =
    \mathcal{M}_{k+1}
    \setminus
    \mathcal{N}_k.
    \label{eq:old_memory}
\end{equation}

\begin{algorithm}[t]
\caption{\textbf{ReTTT: In-Episode Procedural Adaptation}}
\label{alg:rettt}
\footnotesize
\begin{algorithmic}[1]
\Require Task $g$, frozen backbone $\theta$, fast weights $\phi_0$, replay budget $r_{\max}$
\State $\mathcal{M}_0 \gets \emptyset,\; k \gets 0$
\For{$t=1,2,\ldots$ until termination}
    \State $a_t \sim \pi_{\theta,\phi_k}(\cdot \mid g,h_t)$
    \State Execute $a_t$ and update $h_{t+1}$
    \If{$t$ is an adaptation point}
        \State $\mathcal{Z}_k
        \sim
        R_{\theta,\phi_k}
        (\cdot \mid g,h_{t+1},\mathcal{M}_k)$
        
        \State $\widetilde{\mathcal{Z}}_k
        \gets
        \mathrm{Screen}(\mathcal{Z}_k;h_{t+1})$
        
        \State $(\mathcal{M}_{k+1},\mathcal{N}_k)
        \gets
        \mathrm{Maintain}(\mathcal{M}_k,\widetilde{\mathcal{Z}}_k)$
        
        \State $\mathcal{O}_k
        \gets
        \mathcal{M}_{k+1}\setminus\mathcal{N}_k$
        
        \If{$\mathcal{N}_k \neq \emptyset$}
            \State $r_k
            \gets
            \min(r_{\max},|\mathcal{O}_k|)$
            
            \State $\mathcal{R}_k
            \sim
            \mathrm{Sample}(\mathcal{O}_k,r_k)$
            
            \State $\mathcal{D}_k
            \gets
            \mathcal{N}_k \cup \mathcal{R}_k$
            
            \State $\phi_{k+1}
            \gets
            \mathrm{TTT}(\phi_k;\mathcal{D}_k,\theta)$
        \Else
            \State $\phi_{k+1} \gets \phi_k$
        \EndIf
        
        \State $k \gets k+1$
    \EndIf
\EndFor
\end{algorithmic}
\end{algorithm}

\textbf{Memory bounds.}
The procedural memory is constrained by
\begin{equation}
    |\mathcal{M}_{k+1}| \leq B,
    \qquad
    \sum_{m\in\mathcal{M}_{k+1}}\tau(m) \leq T,
    \label{eq:memory_constraints}
\end{equation}
where $B$ is the entry budget, $\tau(m)$ is the number of target tokens associated with entry $m$, $T$ is the total target-token budget.
When either budget is exceeded, generic procedures are evicted before
task-specific ones, with remaining ties resolved by First-In-First-Out (FIFO) order.


\subsection{Test-Time Procedural Internalization}
\label{sec:writeback}

At adaptation event $k$, ReTTT trains on all newly inserted or modified
procedures in $\mathcal{N}_k$ together with a bounded replay subset from
the unchanged memory $\mathcal{O}_k$.
We set
\begin{equation}
r_k
=
\min\left(r_{\max}, |\mathcal{O}_k|\right),
\qquad
\mathcal{R}_k
\sim
\operatorname{Unif}
\left(
\left\{
\mathcal{R}\subseteq\mathcal{O}_k
\;\middle|\;
|\mathcal{R}|=r_k
\right\}
\right),
\label{eq:replay}
\end{equation}
where $r_{\max}$ is the replay budget, $r_k$ is the number of unchanged
procedures replayed at event $k$, and $\mathcal{R}_k$ is sampled uniformly
from all subsets of $\mathcal{O}_k$ of size $r_k$.
We then construct the training set
\begin{equation}
\mathcal{D}_k
=
\mathcal{N}_k \cup \mathcal{R}_k,
\label{eq:training_set}
\end{equation}
so that every newly inserted or modified procedure is used for adaptation,
while only a bounded number of unchanged procedures are replayed.
If $\mathcal{N}_k=\emptyset$, no parameter update is performed.

Each $(x_i,y_i)\in\mathcal{D}_k$ is serialized as a procedural training example, where $x_i$ is the procedural query or checkpoint context and $y_i$ is the corresponding evidence-supported target.
Only target tokens contribute to the test-time objective:
\begin{equation}
    \mathcal{L}_{\mathrm{ReTTT}}(\phi_k)
    =
    -
    \sum_{(x_i,y_i)\in\mathcal{D}_k}
    \sum_{j=1}^{|y_i|}
    \log
    p_{\theta,\phi_k}
    \left(
        y_{i,j}
        \mid
        x_i,y_{i,<j}
    \right).
    \label{eq:rettt_loss}
\end{equation}
The episode-local fast weights are then updated by
\begin{equation}
    \phi_{k+1}
    =
    \phi_k
    -
    \eta
    \nabla_{\phi}
    \mathcal{L}_{\mathrm{ReTTT}}(\phi_k),
    \label{eq:update}
\end{equation}
while the backbone parameters $\theta$ remain frozen.
Raw trajectory tokens are never directly optimized as targets; parameter updates are driven only by the maintained procedural experience in $\mathcal{D}_k$.

These design choices separate experience construction from parameter write-back.
Only screened and maintained procedural experience receives gradient updates, while the explicit memory remains directly revisable across adaptation events.
Because action generation and reflection generation share the same fast weights $\phi_k$, each update changes both subsequent behavior and the policy that constructs later procedural experience.
ReTTT therefore forms a closed in-episode adaptation loop in which maintained procedural experience is progressively internalized into the policy that continues the task.
\section{Experiments}
\label{sec:experiments}

\begin{table}[t]
\centering
\small
\caption{
Task success (\%), averaged over three seeds.
ALFWorld uses 140 tasks, ScienceWorld 120 instances, and SWE-bench Lite 60 instances.
}
\label{tab:main_results}

\setlength{\tabcolsep}{4.5pt}
\renewcommand{\arraystretch}{1.08}

\begin{tabular}{lcccccc}
\toprule
&
\multicolumn{2}{c}{\textbf{ALFWorld}} &
\multicolumn{2}{c}{\textbf{ScienceWorld}} &
\multicolumn{2}{c}{\textbf{SWE-bench Lite}} \\
\cmidrule(lr){2-3}
\cmidrule(lr){4-5}
\cmidrule(lr){6-7}
\textbf{Method}
& \textbf{Qwen}
& \textbf{Gemma}
& \textbf{Qwen}
& \textbf{Gemma}
& \textbf{Qwen}
& \textbf{Gemma} \\
\midrule

\rowcolor{groupgray}
\multicolumn{7}{l}{\textit{Baselines}} \\
Base
& 47.14 & 52.86
& 10.83 & 7.50
& 33.33 & 8.33 \\
qTTT
& 50.00 & 54.29
& 11.67 & 11.67
& 36.67 & 10.00 \\
aTTT
& 51.43 & 53.57
& 15.00 & 10.00
& 40.00 & 11.67 \\

\rowcolor{groupgray}
\multicolumn{7}{l}{\textit{ReTTT}} \\
\rowcolor{retttgreen}
\textbf{ReTTT-QA}
& \textbf{63.57} & 57.86
& \textbf{26.67} & 15.83
& \textbf{43.33} & 18.33 \\
\rowcolor{retttorange}
\textbf{ReTTT-Sum}
& 60.00 & \textbf{59.29}
& 23.33 & \textbf{16.67}
& \textbf{43.33} & \textbf{20.00} \\

\bottomrule
\end{tabular}
\end{table}





\subsection{Experimental Setup}
\label{sec:exp_setup}

\textbf{Benchmarks.}
We evaluate on three long-horizon interactive benchmarks:
ALFWorld~\citep{shridhar2020alfworld} for embodied household tasks,
ScienceWorld~\citep{wang2022scienceworld} for multi-step scientific procedures,
and SWE-bench Lite~\citep{jimenez2024swe} for repository-level software engineering.
We use all 140 ALFWorld tasks, first 120 ScienceWorld instances, and first 60 SWE-bench Lite instances.
The subsets are determined solely by the benchmark ordering and are not selected or filtered based on model performance.

\textbf{Models and adaptation.}
We evaluate Qwen3.5-9B~\citep{qwen3.5} and Gemma4-12B~\citep{team2026gemma}.
The backbone remains frozen within each episode, and test-time updates are restricted to lightweight LoRA fast weights shared by action and reflection generation.

\textbf{Baselines.}
We compare against Base, the frozen agent without parameter updates;
qTTT~\citep{bansal2026let}, which performs query-only test-time training;
and aTTT~\citep{wang2026no}, which continually trains on agent trajectories.
Our two variants, ReTTT-QA and ReTTT-Sum, differ in whether procedural experience is serialized as QA-style lessons or structured progress summaries.

\textbf{Implementation details.}
We use a procedural-memory capacity of 12, reflect every 5 interaction steps, replay up to 2 previous procedures per update, and learning rate is $5\times 10^{-4}$.
Additional implementation details are provided in Appendix~\ref{app:implementation}.

\subsection{Main Results}
\label{sec:main_results}

Table~\ref{tab:main_results} reports task success averaged over three seeds.
Across all three benchmarks and both backbones, both ReTTT variants outperform aTTT, and the best ReTTT variant achieves the strongest result in every model--benchmark setting.

\begin{table}[t]
\centering
\small
\caption{
ALFWorld horizon-stratified results with Qwen3.5-9B.
Each cell reports \textbf{success (\%) / steps}.
}
\label{tab:alfworld_horizon}

\setlength{\tabcolsep}{5pt}
\renewcommand{\arraystretch}{1.08}

\begin{tabular}{lcccc}
\toprule
\textbf{Method} & \textbf{Short} & \textbf{Medium} & \textbf{Long} & \textbf{Overall} \\
\midrule
\rowcolor{groupgray}
\multicolumn{5}{l}{\textit{Baselines}} \\
Base
& 72.22 / 20.50
& 51.06 / 30.87
& 38.67 / 39.52
& 47.14 / 34.17 \\
qTTT
& 83.33 / 15.72
& 59.57 / 29.43
& 36.00 / 39.93
& 50.00 / 33.29 \\
aTTT
& 88.89 / 17.33
& 57.45 / 28.47
& 38.67 / 40.20
& 51.43 / 33.32 \\

\rowcolor{groupgray}
\multicolumn{5}{l}{\textit{ReTTT}} \\
\rowcolor{retttgreen}
\textbf{ReTTT-QA}
& 94.44 / 13.06
& \textbf{68.09} / \textbf{26.57}
& \textbf{53.33} / \textbf{33.92}
& \textbf{63.57} / \textbf{28.77} \\
\rowcolor{retttorange}
\textbf{ReTTT-Sum}
& \textbf{100.00} / \textbf{9.33}
& 65.96 / 27.36
& 46.67 / 36.69
& 60.00 / 30.04 \\
\bottomrule
\end{tabular}
\end{table}

\textbf{ALFWorld.}
With Qwen3.5-9B, ReTTT-QA improves success from 51.43\% with aTTT to 63.57\%; with Gemma4-12B, ReTTT-Sum improves from 53.57\% to 59.29\%.

We further stratify ALFWorld tasks by reference low-level actions $L^\star$
in Table~\ref{tab:alfworld_horizon}, defining Short, Medium, and Long tasks as
$L^\star\leq25$, $26\leq L^\star\leq40$, and $L^\star\geq41$, respectively.
For ReTTT-QA, the gain over aTTT increases from 5.6 points on Short tasks to 10.6 on Medium and 14.7 on Long tasks.
ReTTT is particularly effective on more demanding multi-stage tasks.

\begin{wraptable}{r}{0.52\linewidth}
\vspace{-0.8em}
\centering
\small
\caption{
ALFWorld execution statistics with Qwen3.5-9B.
Rates are normalized by total average executed steps over three seeds.
}
\label{tab:alfworld_execution}

\setlength{\tabcolsep}{3pt}
\renewcommand{\arraystretch}{1.07}

\begin{tabular}{lccc}
\toprule
\textbf{Method}
& \textbf{Total Steps}
& \textbf{Unparseable}
& \textbf{Inadmissible} \\
\midrule

Base
& 4784
& 65 (1.36\%)
& 860 (17.98\%) \\
qTTT
& 4661
& 91 (1.95\%)
& 686 (14.72\%) \\
aTTT
& 4665
& 107 (2.29\%)
& 913 (19.57\%) \\

\midrule

\rowcolor{retttgreen}
\textbf{ReTTT-QA}
& \textbf{4028}
& \textbf{40 (0.99\%)}
& 617 (15.32\%) \\
\rowcolor{retttorange}
\textbf{ReTTT-Sum}
& 4206
& 50 (1.19\%)
& \textbf{552 (13.12\%)} \\

\bottomrule
\end{tabular}

\vspace{-0.8em}
\end{wraptable}

The gains are accompanied by shorter and more stable execution.
On Long tasks, ReTTT-QA and ReTTT-Sum use 33.92 and 36.69 steps on average, respectively, compared with 40.20 for aTTT, while both reach higher success versus aTTT.
Table~\ref{tab:alfworld_execution} further shows that ReTTT-QA reduces total interaction steps from 4665 to 4028 and lowers the unparseable- and inadmissible-action rates from 2.29\% and 19.57\% to 0.99\% and 15.32\%.
ReTTT-Sum further lowers the inadmissible-action rate to 13.12\%.

\textbf{ScienceWorld and SWE-bench Lite.}
The gains extend to substantially different interaction domains.
On ScienceWorld, the best ReTTT results reach 26.67\% with Qwen3.5-9B and 16.67\% with Gemma4-12B, compared with 15.00\% and 10.00\% for aTTT.
On SWE-bench Lite, ReTTT reaches 43.33\% with Qwen3.5-9B and 20.00\% with Gemma4-12B, compared with 40.00\% and 11.67\% for aTTT.
These results indicate that the benefit of procedural adaptation extends beyond embodied state tracking to scientific interaction and software engineering.

\subsection{Is Parametric Procedural Memory More Effective than Explicit Context Memory?}
\label{sec:parametric_memory}

We first ask whether extracted procedural experience is better retained explicitly in context than internalized into model parameters.
To isolate this factor, we construct Context Memory, which uses the same procedural reflection and memory pipeline as ReTTT but keeps the model frozen and supplies the maintained experience only as textual context.

As shown in Figure~\ref{fig:parametric_memory}, Context Memory improves success from 47.14\% to 53.57\%, showing that explicit procedural memory is already useful.
Parametric internalization yields substantially larger gains: ReTTT-Sum reaches 60.00\% and ReTTT-QA 63.57\%, with ReTTT-QA outperforming Context Memory by 10.00 points.
Thus, procedural experience is more effective when internalized into fast weights than when reused only as explicit context.

\subsection{Is Procedural TTT Effective without Reflection-Policy Adaptation?}
\label{sec:closed_loop_exp}

We next isolate the role of adapting the reflection policy.
Frozen Reflection retains ReTTT's procedural reflection, memory maintenance, and fast-weight updates for action generation, but generates every reflection with the initial policy $\phi_0$ rather than the evolving $\phi_k$.

\begin{figure*}[t]
    \centering

    \begin{subfigure}[t]{0.455\textwidth}
        \centering
        \begin{minipage}[t][4.2cm][t]{\linewidth}
            \centering
            \vspace{0pt}
            \includegraphics[width=\linewidth]
            {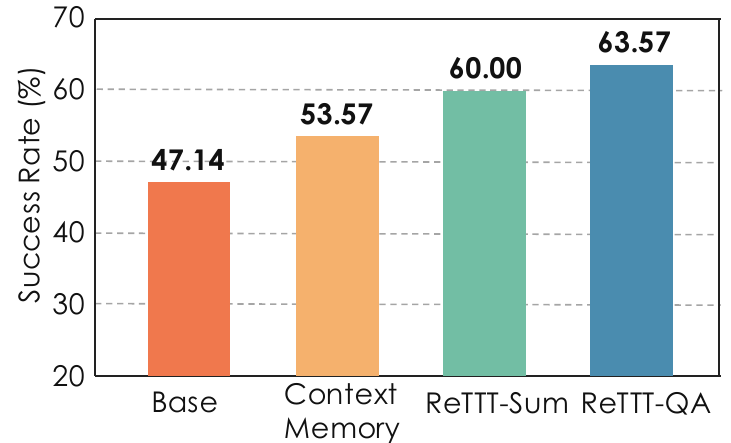}
        \end{minipage}
        \caption{Explicit context memory vs.\ parametric internalization.}
        \label{fig:parametric_memory}
    \end{subfigure}
    \hfill
    \begin{subfigure}[t]{0.51\textwidth}
        \centering
        \begin{minipage}[t][4.2cm][t]{\linewidth}
            \centering
            \vspace{0pt}
            \includegraphics[width=\linewidth]
            {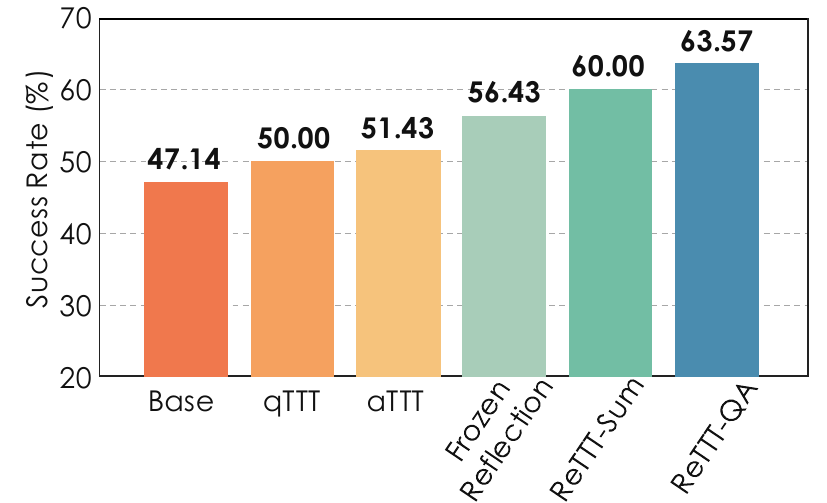}
        \end{minipage}
        \caption{Procedural TTT with frozen vs.\ adapted reflection.}
        \label{fig:frozen_reflection}
    \end{subfigure}

    \caption{
        \textbf{Two controlled comparisons on ALFWorld with Qwen3.5-9B.}
    }
    \label{fig:internalization_and_reflection}
\end{figure*}

As shown in Figure~\ref{fig:frozen_reflection}, Frozen Reflection reaches 56.43\%, outperforming qTTT 50.00\% and aTTT 51.43\%.
Thus, maintained procedural supervision remains effective even with a fixed reflector.
Allowing reflection to share the evolving fast weights further raises ReTTT-QA to 63.57\%, a 7.14-point gain, showing an additional benefit from adapting the policy that constructs future procedural experience.

\begin{figure}[t]
    \centering
    \includegraphics[width=0.99\textwidth]{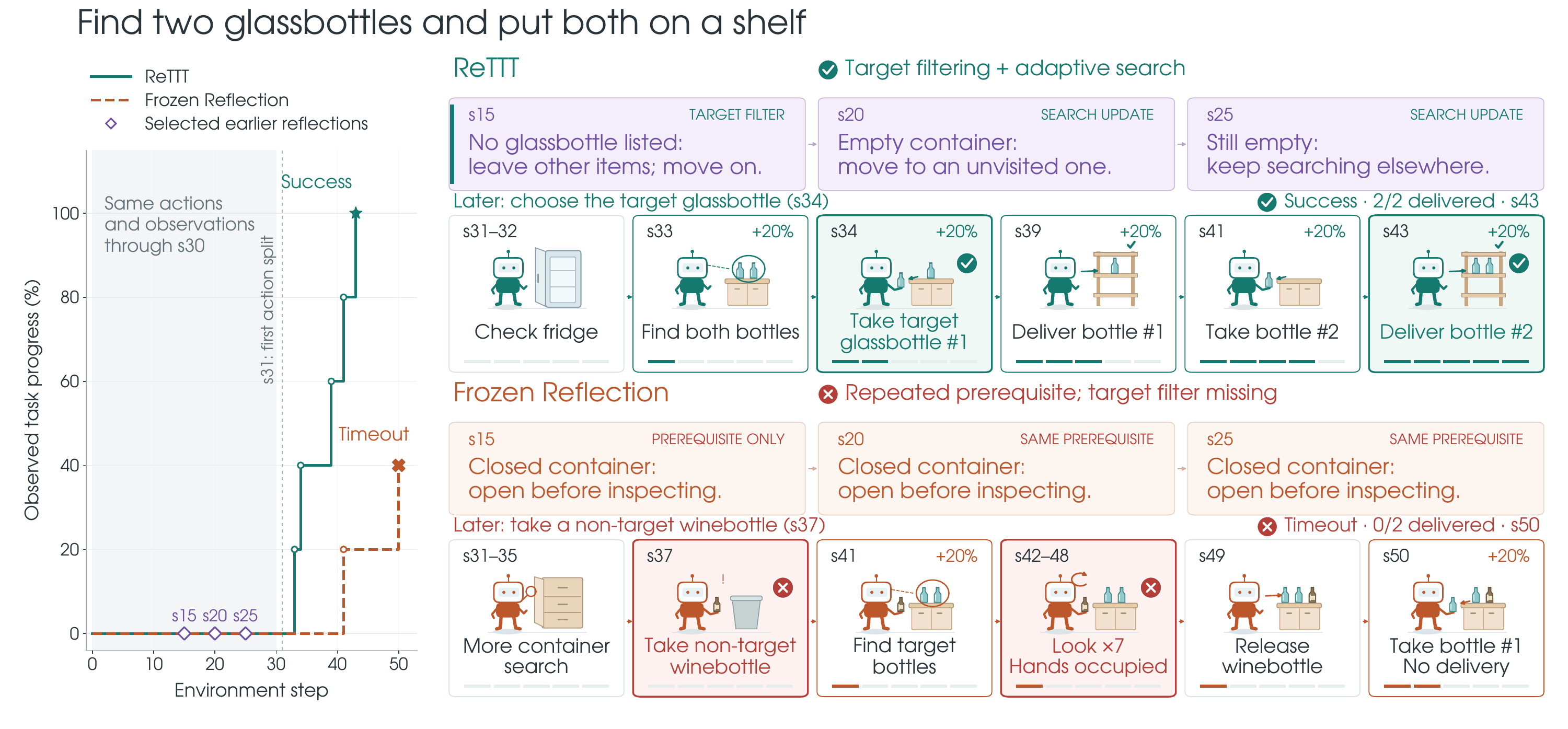}
    \caption{\textbf{Trajectory-level comparison of ReTTT and Frozen Reflection.}}
    \label{fig:reflection_trajectory}
\end{figure}

\textbf{Trajectory-level behavior.}
Figure~\ref{fig:reflection_trajectory} provides a controlled example: ReTTT and Frozen Reflection
share the same trajectory through step 30, but the adapted reflector turns failed
searches into task-specific exploration guidance, whereas the frozen reflector
repeats generic advice.
ReTTT subsequently completes both deliveries by step 43, while Frozen Reflection times out at step 50.


\subsection{Ablation Study}
\label{sec:ablation}

We ablate three components of ReTTT-QA, as shown in Table~\ref{tab:ablation}:
candidate screening, dynamic memory maintenance, and replay.

\begin{wraptable}{r}{0.47\linewidth}
\vspace{-0.7em}
\centering
\small
\caption{Ablation study on ALFWorld with Qwen3.5-9B.}
\label{tab:ablation}

\setlength{\tabcolsep}{5pt}
\renewcommand{\arraystretch}{1.08}

\begin{tabular}{lc}
\toprule
\textbf{Variant}
& \textbf{Success (\%)} \\
\midrule

\rowcolor{retttgreen}
\textbf{ReTTT-QA}
& \textbf{63.57} \\

\quad w/o Replay
& 62.14 \\
\quad w/o Candidate Screening
& 59.29 \\
\quad w/o Dynamic Maintenance
& 57.86 \\

\bottomrule
\end{tabular}

\vspace{-0.8em}
\end{wraptable}

\textbf{Candidate screening.}
Removing screening causes the drop from 63.57\% to 59.29\%.
Without evidence- and trajectory-based validation, unsupported or inconsistent procedures can enter the training set and degrade the adaptation signal.

\textbf{Dynamic memory maintenance.}
Disabling dynamic maintenance reduces success to 57.86\%.
Without similarity-based replacement and revision, redundant or outdated procedures can persist as new evidence arrives. It has a significant impact on performance.

\textbf{Replay.}
Removing replay yields a smaller decrease to 62.14\%.
Training only on newly inserted or revised procedures suggests that replay mainly stabilizes continual adaptation by revisiting previously retained experience.

Overall, these results indicate that the quality and consistency of procedural supervision are more important than simply increasing repeated training.
Additional efficiency analysis is provided in Appendix~\ref{sec:efficiency}.

\section{Related Work}
\label{sec:related}

\subsection{Memory for Language Agents}

Language-agent memory has evolved from explicit storage toward increasingly structured forms of experience reuse.
MemGPT~\citep{packer2023memgpt} manages external memory beyond the immediate context, while Generative Agents~\citep{park2023generative} and Reflexion~\citep{shinn2023reflexion} retain episodic experience and reflection in textual memory.
Voyager~\citep{wang2023voyager} further stores reusable executable skills, shifting memory from recalling past observations toward preserving procedural knowledge.

Recent work makes such experience more adaptive.
AdaMEM~\citep{zhang2026adamem} dynamically constructs test-time strategy memory, and ProcMEM~\citep{mi2026procmem} learns and maintains reusable procedural skills from interaction, both without online parameter updates.
Complementary approaches begin to parameterize experience: ParamMem~\citep{yao2026parammem} encodes cross-sample reflection patterns into parametric memory, while Procedural Memory Distillation~\citep{liu2026procedural} distills accumulated procedural memory into model weights during training.
ReTTT instead targets in-episode procedural internalization: interaction experience is continually reflected, maintained as explicit procedural memory, and written into episode-local fast weights while the same task is still unfolding.

\subsection{Test-Time Adaptation for Language Agents}

Test-time training adapts model parameters using information available only at inference time.
Early TTT methods~\citep{sun2020test} establish this paradigm for distribution adaptation, while qTTT~\citep{bansal2026let} shows that context-specific gradient updates can improve long-context utilization in LLMs.
These settings primarily adapt to a fixed test input or context.
More recent work extends parameter adaptation to interactive agents.
Agentic TTT (aTTT)~\citep{wang2026no} continually trains on the evolving agent trajectory and reweights repeated tokens to mitigate self-training drift.

ReTTT differs in how reflection, memory, and adaptation are coupled and the currently adapted policy first converts interaction history into screened and dynamically maintained procedural experience.
This experience is then internalized into fast weights shared by both action and reflection generation, so earlier updates influence not only subsequent behavior but also how later experience is interpreted and transformed into future supervision.
\section{Conclusion}
\label{sec:conclusion}

We introduced ReTTT, which constructs and internalizes maintained procedural
experience for in-episode test-time adaptation, with shared fast weights coupling
action and reflection in a closed loop.
Across three benchmarks and two model families, ReTTT consistently outperforms
trajectory-based agentic TTT, with larger gains on longer-horizon tasks,
highlighting the value of reliable procedural supervision and adaptive reflection.
Future work can extend ReTTT across episodes by selectively consolidating and
transferring reusable procedural experience.

\bibliography{iclr2027_conference}
\bibliographystyle{iclr2027_conference}

\newpage
\appendix
\section{Appendix}


\subsection{Test-Time Compute and Efficiency}
\label{sec:efficiency}

ReTTT introduces additional generation cost through reflection while using fewer optimization steps than existing test-time training baselines.
We characterize this trade-off using reflection tokens, loss-bearing target-token exposures, adapter gradient steps, environment interaction, and aggregate worker time.
Table~\ref{tab:efficiency} reports per-episode statistics for the Qwen3.5-9B ALFWorld runs.

ReTTT changes the allocation of test-time computation.
ReTTT-QA uses 524.78 loss-bearing target exposures and 7.14 gradient steps per episode, compared with 573.69 exposures and 12.93 gradient steps for aTTT, while improving success from 51.43\% to 63.57\%.
ReTTT-Sum further reduces the optimization load to 257.99 target exposures and 2.51 gradient steps.
Both ReTTT variants also require substantially fewer gradient steps than qTTT.

The reduced optimization load is accompanied by shorter environment interaction.
As shown in Table~\ref{tab:alfworld_horizon}, ReTTT-QA reduces the average rollout length from 33.32 steps with aTTT to 28.77 steps overall, and from 40.20 to 33.92 steps on long-horizon tasks.
Table~\ref{tab:alfworld_execution} further shows that ReTTT-QA reduces total executed steps from 4665 to 4028 while producing fewer malformed and inadmissible actions.
Thus, ReTTT achieves higher task success with fewer gradient updates and shorter interaction trajectories.

This reduction in optimization does not make ReTTT cheaper overall, because reflection introduces additional generation cost.
The aggregate worker time of ReTTT-QA and ReTTT-Sum is approximately 1.085$\times$ and 1.124$\times$ that of aTTT, respectively.
ReTTT therefore shifts test-time computation away from repeated optimization and toward the construction of procedural supervision, achieving stronger performance with a modest increase in aggregate runtime.

\begin{table}[ht]
\centering
\small
\caption{
Test-time computation on ALFWorld with Qwen3.5-9B.
\emph{Refl. In / Out} denotes average reflection input/output tokens per episode.
\emph{Target Exp.} counts loss-bearing target-token exposures across optimization steps.
\emph{Worker Time} is summed worker runtime normalized by Base.
}
\label{tab:efficiency}

\setlength{\tabcolsep}{4.5pt}
\renewcommand{\arraystretch}{1.08}

\begin{tabular}{lccccc}
\toprule
\textbf{Method}
& \textbf{SR}
& \textbf{Refl. In / Out}
& \textbf{Target Exp.}
& \textbf{Grad. Steps}
& \textbf{Worker Time($\times$)} \\
\midrule

\rowcolor{groupgray}
\multicolumn{6}{l}{\textit{Baselines}} \\
Base
& 47.14
& 0 / 0
& 0
& 0
& 1.000 \\
qTTT
& 50.00
& 0 / 0
& 4064.00
& 32.00
& 2.001 \\
aTTT
& 51.43
& 0 / 0
& 573.69
& 12.93
& 2.724 \\

\rowcolor{groupgray}
\multicolumn{6}{l}{\textit{ReTTT}} \\
\rowcolor{retttgreen}
\textbf{ReTTT-QA}
& \textbf{63.57}
& 27875.66 / 1175.30
& 524.78
& 7.14
& 2.955 \\
\rowcolor{retttorange}
\textbf{ReTTT-Sum}
& 60.00
& 29442.70 / 1001.23
& \textbf{257.99}
& \textbf{2.51}
& 3.061 \\

\bottomrule
\end{tabular}
\end{table}

\subsection{Implementation Details and Hyperparameters}
\label{app:implementation}

We use the same test-time adaptation configuration
across benchmarks and model backbones.
The backbone parameters remain frozen throughout each episode, and only
episode-local LoRA fast weights are optimized.
The adapter, optimizer state, and procedural memory are reset after each
episode.
Table~\ref{tab:shared_hyperparameters} summarizes the common adaptation
hyperparameters.

\begin{table}[ht]
\centering
\small
\caption{Shared test-time adaptation hyperparameters.}
\label{tab:shared_hyperparameters}
\setlength{\tabcolsep}{6pt}
\renewcommand{\arraystretch}{1.08}

\begin{tabular}{ll}
\toprule
\textbf{Setting} & \textbf{Value} \\
\midrule

Backbone precision
& bfloat16 \\

Adapter precision
& float32 \\

Adapted module
& Final attention \texttt{q\_proj} \\

Parallel LoRA components
& 2 \\

LoRA rank
& 4 \\

LoRA $\alpha$
& 8 \\

Optimizer
& AdamW \\

Learning rate
& $5\times10^{-4}$ \\

AdamW $(\beta_1,\beta_2)$
& $(0.9,0.999)$ \\

AdamW $\epsilon$
& $10^{-8}$ \\

Weight decay
& 0 \\

Gradient clipping
& 1.0 \\

Maximum training length
& 1,024 tokens \\

Supervised tokens
& Target \\

Memory capacity
& 12 entries \\

Memory target-token budget
& 768 tokens \\

Cross-episode persistence
& None \\

\bottomrule
\end{tabular}
\end{table}

Reflection is performed periodically during interaction using the same
episode-local adapted model as action generation.
Generated procedural experience is screened and maintained before being used
for parameter updates.
ReTTT-QA and ReTTT-Sum share the same adaptation mechanism but differ in the
serialization of procedural experience, as summarized
in Table~\ref{tab:variant_hyperparameters}.

\begin{table}[ht]
\centering
\small
\caption{Variant-specific settings of ReTTT-QA and ReTTT-Sum.}
\label{tab:variant_hyperparameters}
\setlength{\tabcolsep}{6pt}
\renewcommand{\arraystretch}{1.08}

\begin{tabular}{lcc}
\toprule
\textbf{Setting}
& \textbf{ReTTT-QA}
& \textbf{ReTTT-Sum} \\
\midrule

Representation
& Procedural QA
& Progress summary \\

Generated items / reflection
& $\leq 3$
& $\leq 1$ \\

Maximum example updates
& 5
& 3 \\

Memory update
& Item-level
& Latest checkpoint \\

\bottomrule
\end{tabular}
\end{table}

\subsection{Additional Details of Procedural Memory Maintenance}
\label{app:memory_maintenance}

We provide additional implementation details of the candidate screening,
memory reconciliation, revision, and eviction operations used in ReTTT.
Each procedural candidate is represented as
$z=(\kappa,u,E)$, where $\kappa$ denotes its procedural type, $u$ its
decision-relevant content, and $E$ its supporting trajectory evidence.
All operations described below are deterministic and introduce no additional
model calls beyond the reflection step described in the main method.

\paragraph{Candidate screening.}
Before memory reconciliation, each reflected candidate is passed through a
deterministic screening stage. We first verify that the candidate follows the
expected procedural schema, that the cited evidence refers to valid steps in
the current interaction trajectory, and that the serialized training target
satisfies the prescribed token budget.

Evidence consistency is checked using fixed rules and string-matching
predicates over the cited trajectory observations, environment feedback, and
task outcome. In particular, a candidate is not accepted merely because its
evidence identifiers are valid: procedural claims that contradict the
corresponding trajectory evidence are rejected. For example, claims about
successful progress must be supported by the associated environment feedback
rather than only by an attempted action. Screening therefore performs
rule-based evidence grounding and trajectory-consistency checks, without an
additional LLM judge, learned verifier, or entailment model. Candidates that
fail any screening condition are discarded before entering memory or
contributing to parameter updates.

\paragraph{Similarity-based reconciliation.}
Each surviving candidate is compared only with memory entries of the same
procedural type. Entries with a different procedural type are not considered
as reconciliation candidates. For two same-type entries $z$ and $m$, we
compute Jaccard similarity over their normalized decision-relevant content:
\begin{equation}
    s(z,m)
    =
    \frac{
        \left|
        C(z) \cap C(m)
        \right|
    }{
        \left|
        C(z) \cup C(m)
        \right|
    },
\end{equation}
where $C(\cdot)$ denotes the normalized content representation used for
memory matching. We then select
\begin{equation}
    m^*
    =
    \operatorname*{arg\,max}_{m\in\mathcal{M}_k:\kappa_m=\kappa_z}
    s(z,m),
    \qquad
    s^*=s(z,m^*).
\end{equation}
If no entry with the same procedural type exists, the candidate is directly
treated as a new entry.

We use fixed thresholds
\begin{equation}
    \tau_{\mathrm{low}} = 0.3,
    \qquad
    \tau_{\mathrm{high}} = 0.7,
\end{equation}
and reconcile the candidate according to
\begin{equation}
\operatorname{Reconcile}(z,m^*) =
\begin{cases}
    \operatorname{Insert}(z),
        & s^* < 0.3, \\[1mm]
    \operatorname{Revise}(m^*,z),
        & 0.3 \leq s^* < 0.7, \\[1mm]
    \operatorname{Replace}(m^*,z),
        & s^* \geq 0.7.
\end{cases}
\label{eq:memory_reconciliation}
\end{equation}
Thus, low-overlap candidates introduce new procedures, highly overlapping
candidates replace an existing entry with the newly observed version, and
partially overlapping candidates trigger revision.

\paragraph{Template-based revision.}
For moderate similarity, ReTTT revises the matched memory entry using a
fixed deterministic template rather than an additional language-model call.
The template concatenates the existing procedural content with the newly
observed information into a single updated entry:
\begin{equation}
    m'
    =
    \operatorname{TemplateMerge}(m^*,z).
\end{equation}
This operation is purely rule based and does not invoke the reflection model
or any separate verifier.
The revised entry is included
in $\mathcal{N}_k$ and is therefore eligible for subsequent test-time
internalization.

\paragraph{Memory bounds and eviction.}
After reconciliation, ReTTT enforces both an entry-count budget and a
target-token budget:
\begin{equation}
    |\mathcal{M}_{k+1}| \leq B,
    \qquad
    \sum_{m\in\mathcal{M}_{k+1}}\tau(m) \leq T,
\end{equation}
where $\tau(m)$ denotes the number of target tokens associated with entry
$m$. In our experiments, $B=12$ and $T=768$.

To determine eviction priority, we use a fixed lexical classifier over the
normalized procedural content. Each memory entry is assigned to one of seven
categories shown in Table~\ref{tab:memory_specificity}. The first four
categories capture task-specific progress, requirements, or recovery
information and are therefore retained with higher priority. The remaining
three categories represent more generic procedural guidance and are evicted
first when the memory budget is exceeded. The classifier is rule based and
does not invoke an LLM.

\begin{table}[t]
\centering
\small
\caption{
\textbf{Rule-based specificity classes used for memory eviction.}
Task-specific entries are retained before generic entries.
}
\label{tab:memory_specificity}
\begin{tabular}{ll}
\toprule
Category & Eviction class \\
\midrule
Failure recovery     & Task-specific \\
Multi-object         & Task-specific \\
Task transform       & Task-specific \\
Termination state    & Task-specific \\
\midrule
Generic container    & Generic \\
Generic navigation   & Generic \\
Other                & Generic \\
\bottomrule
\end{tabular}
\end{table}

When either memory constraint is violated, ReTTT first considers entries in
the generic classes for eviction. If multiple generic entries are available,
the oldest one is removed. If no generic entry remains, the oldest
task-specific entry is removed instead. Eviction is repeated until both
memory constraints are satisfied.

FIFO age is determined by the original insertion time of an entry.
Replacing or revising an existing entry does not reset its age for eviction
purposes. Thus, reconciliation can update the content of a stored procedure
without moving it to the end of the FIFO queue.

\subsection{More Case Studies}
\label{app:more_case_studies}

We provide additional trajectory-level examples to complement the controlled
case study in the main text. Across ALFWorld, ScienceWorld, and SWE-bench Lite,
these examples illustrate how ReTTT converts interaction feedback into
task-specific procedural guidance and how the resulting adaptation shapes
subsequent execution.

\begin{figure}[p]
    \centering

    \includegraphics[
        width=0.99\textwidth,
        height=0.38\textheight,
        keepaspectratio
    ]{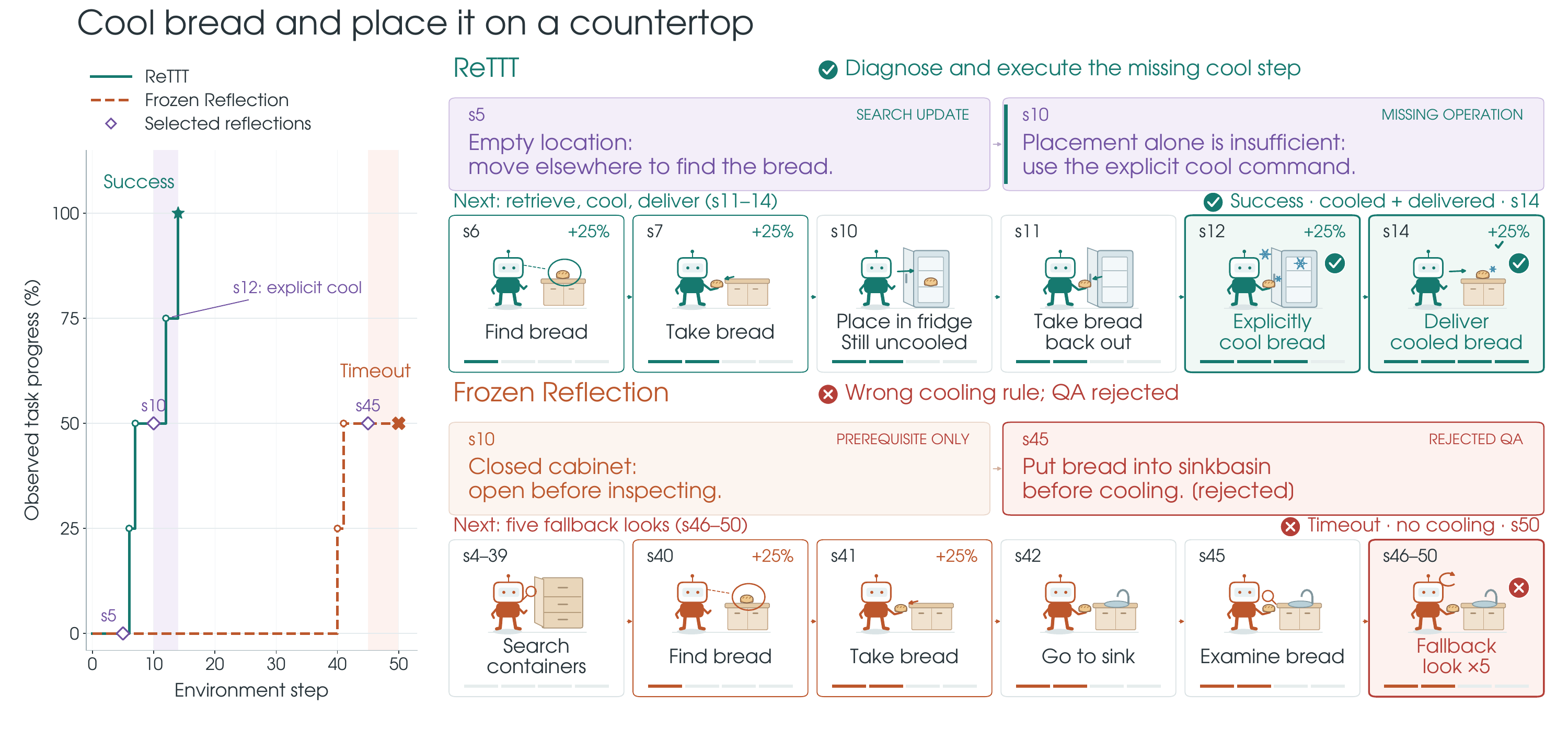}
    \caption{
    \textbf{ALFWorld case: identifying a missing operation.}
    ReTTT recognizes that placement alone does not cool the bread and executes
    the required cooling operation, while Frozen Reflection times out.
    }
    \label{fig:case_alf_cool}

    \vspace{0.6em}

    \includegraphics[
        width=0.99\textwidth,
        height=0.38\textheight,
        keepaspectratio
    ]{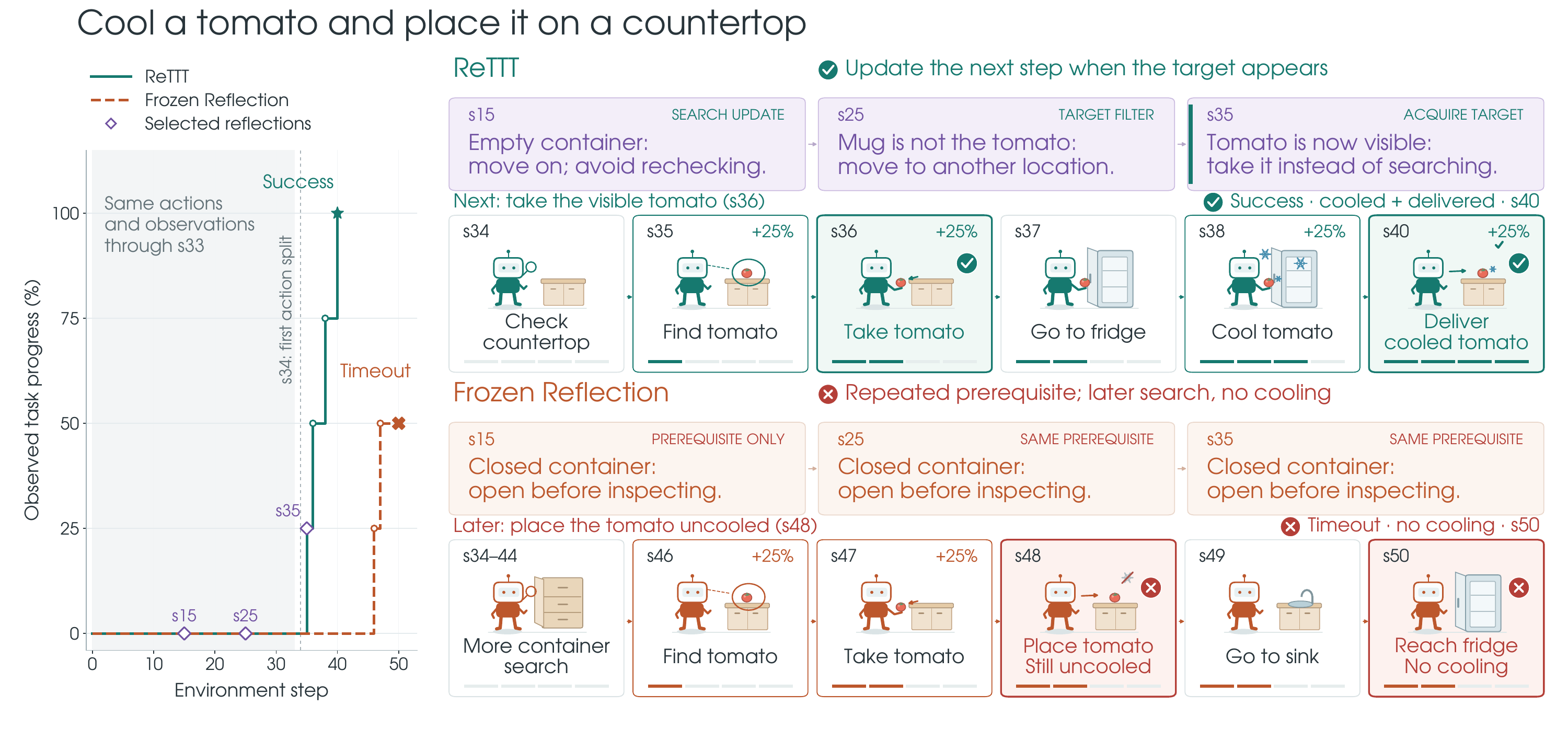}
    \caption{
    \textbf{ALFWorld case: updating the next action from new evidence.}
    ReTTT switches from exploration to target acquisition once the tomato is
    observed, whereas Frozen Reflection continues repetitive guidance and times out.
    }
    \label{fig:case_alf_tomato}
\end{figure}

\paragraph{Additional ALFWorld cases.}
Figures~\ref{fig:case_alf_cool} and~\ref{fig:case_alf_tomato} show the same
qualitative pattern as the example in the main text. In the cooling task,
ReTTT identifies the missing operation---placing the object in the refrigerator
does not itself cool it---and converts this feedback into the explicit action
needed to complete the task. In the second case, once the target tomato becomes
visible, the adapted reflector updates its guidance from continued exploration
to immediate target acquisition. By contrast, Frozen Reflection continues to
produce generic prerequisite-oriented guidance or unsupported rules, leading
to unnecessary search and eventual timeout. Together, these cases show that
adapted reflection can revise procedural guidance as new task-specific evidence
becomes available.

\begin{figure}[p]
    \centering

    \includegraphics[
        width=0.99\textwidth,
        height=0.38\textheight,
        keepaspectratio
    ]{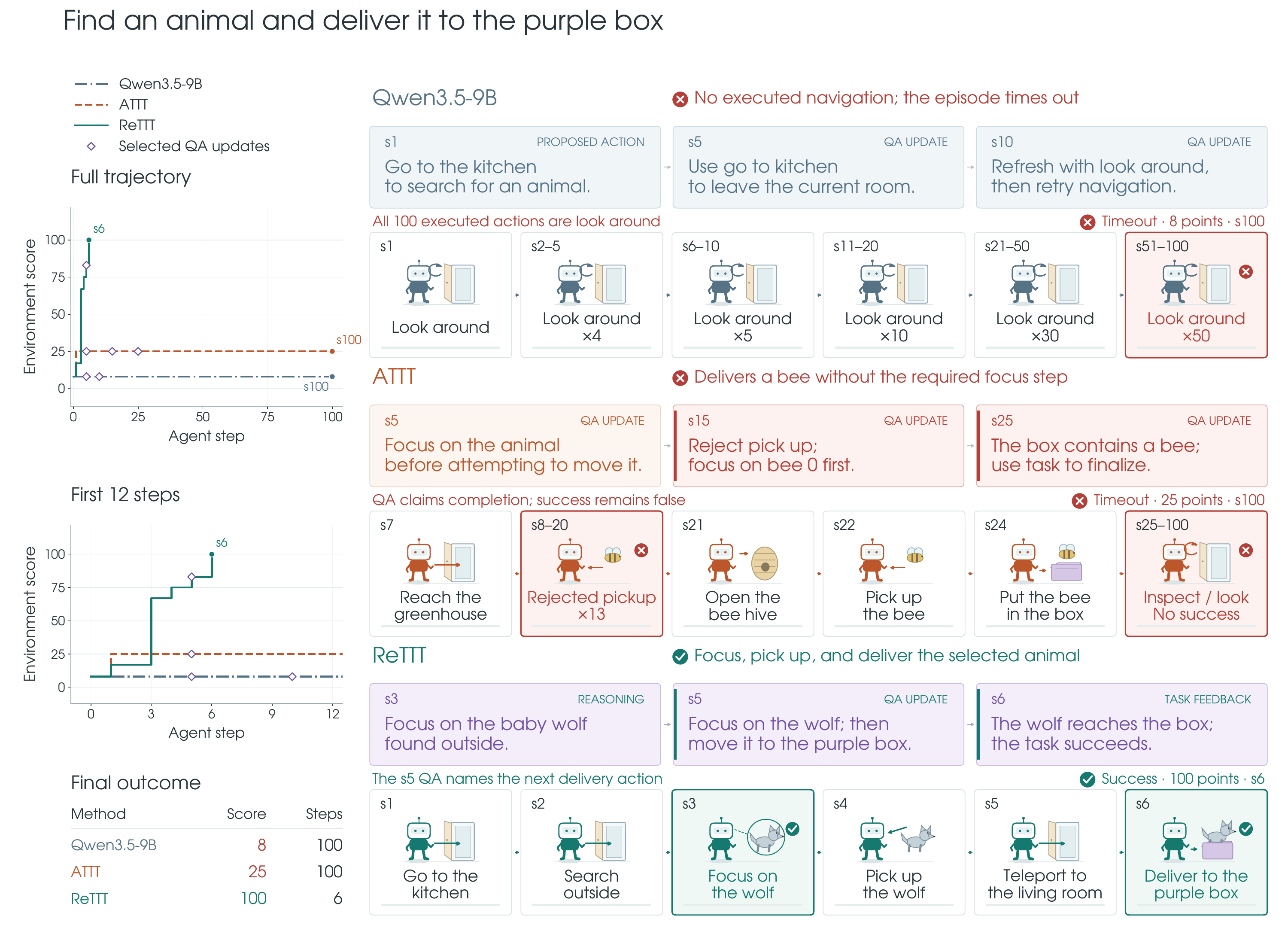}
    \caption{
    \textbf{ScienceWorld case: satisfying a required interaction prerequisite.}
    ReTTT focuses on the selected animal before pickup and delivery, completing
    the task in six steps, while the baselines fail to execute the required sequence.
    }
    \label{fig:case_sci_animal}

    \vspace{0.6em}

    \includegraphics[
        width=0.99\textwidth,
        height=0.38\textheight,
        keepaspectratio
    ]{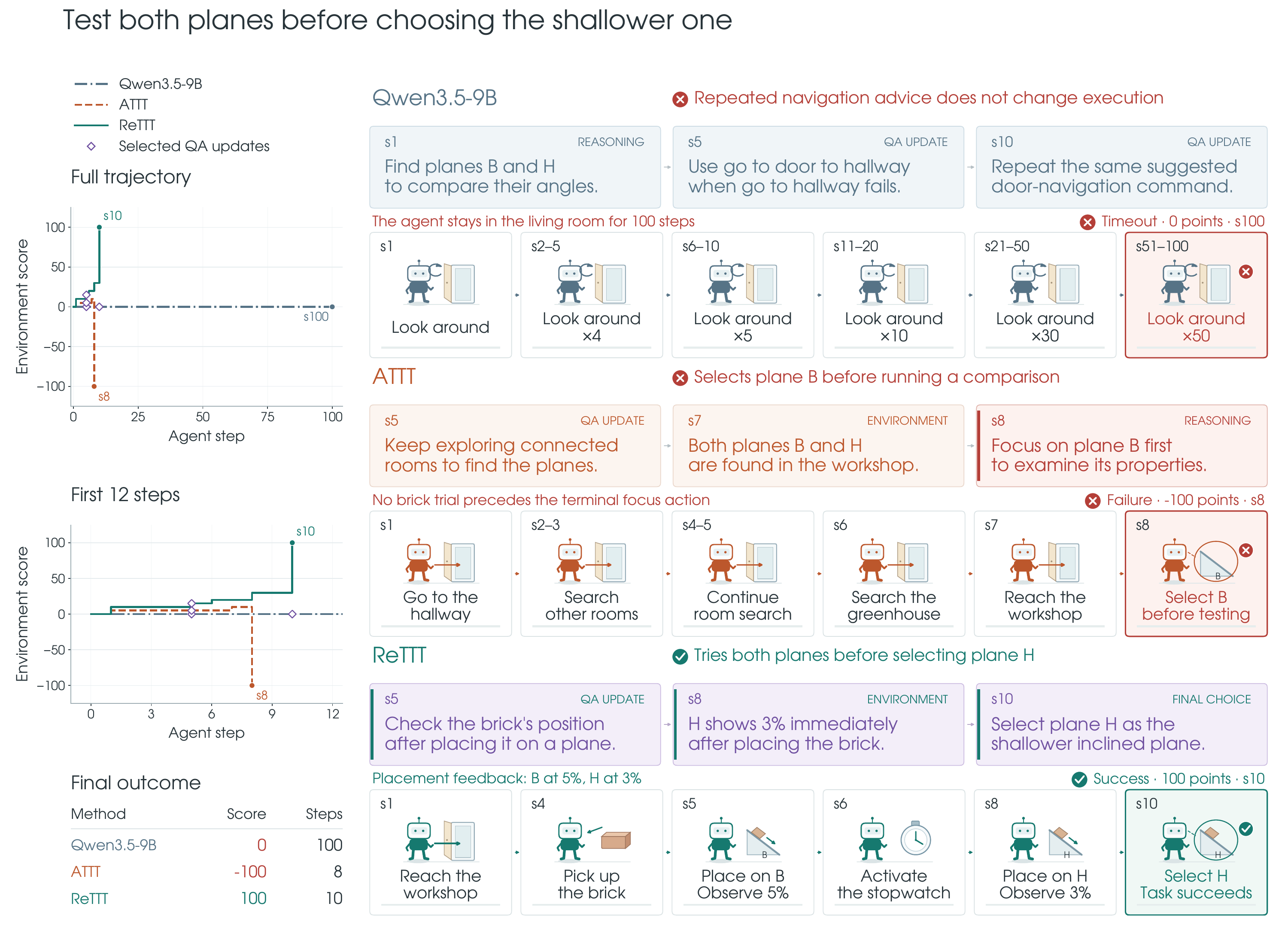}
    \caption{
    \textbf{ScienceWorld case: gathering evidence before selection.}
    ReTTT tests both planes and uses the observed responses to select the
    shallower one, whereas aTTT commits before performing the required comparison.
    }
    \label{fig:case_sci_planes}
\end{figure}

\begin{figure}[p]
    \centering

    \includegraphics[
        width=0.99\textwidth,
        height=0.38\textheight,
        keepaspectratio
    ]{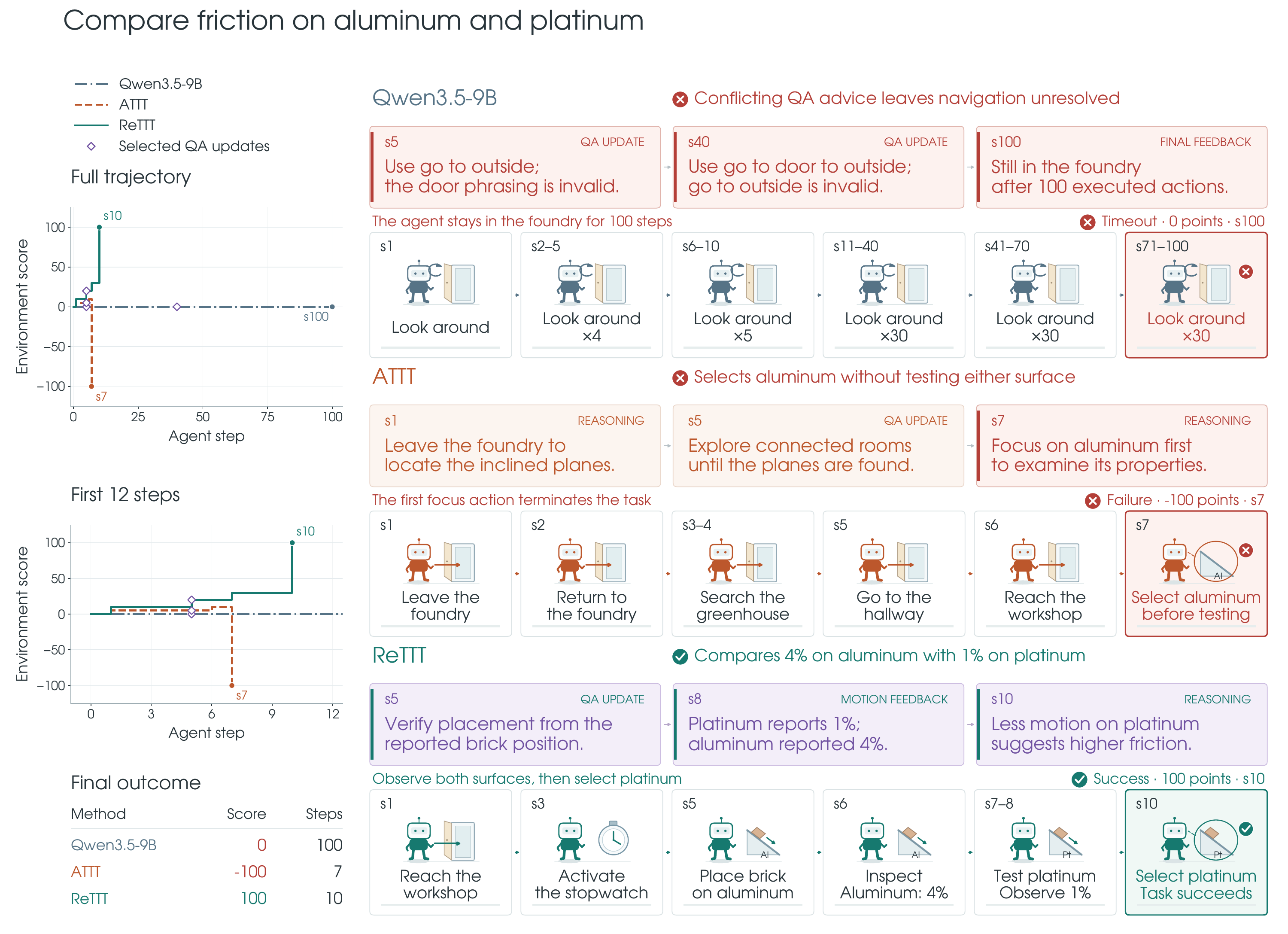}
    \caption{
    \textbf{ScienceWorld case: comparing candidates before committing.}
    ReTTT tests both surfaces and selects platinum from the observed motion
    feedback, while aTTT makes a premature selection without performing the comparison.
    }
    \label{fig:case_sci_friction}

    \vspace{0.6em}

    \includegraphics[
        width=0.99\textwidth,
        height=0.38\textheight,
        keepaspectratio
    ]{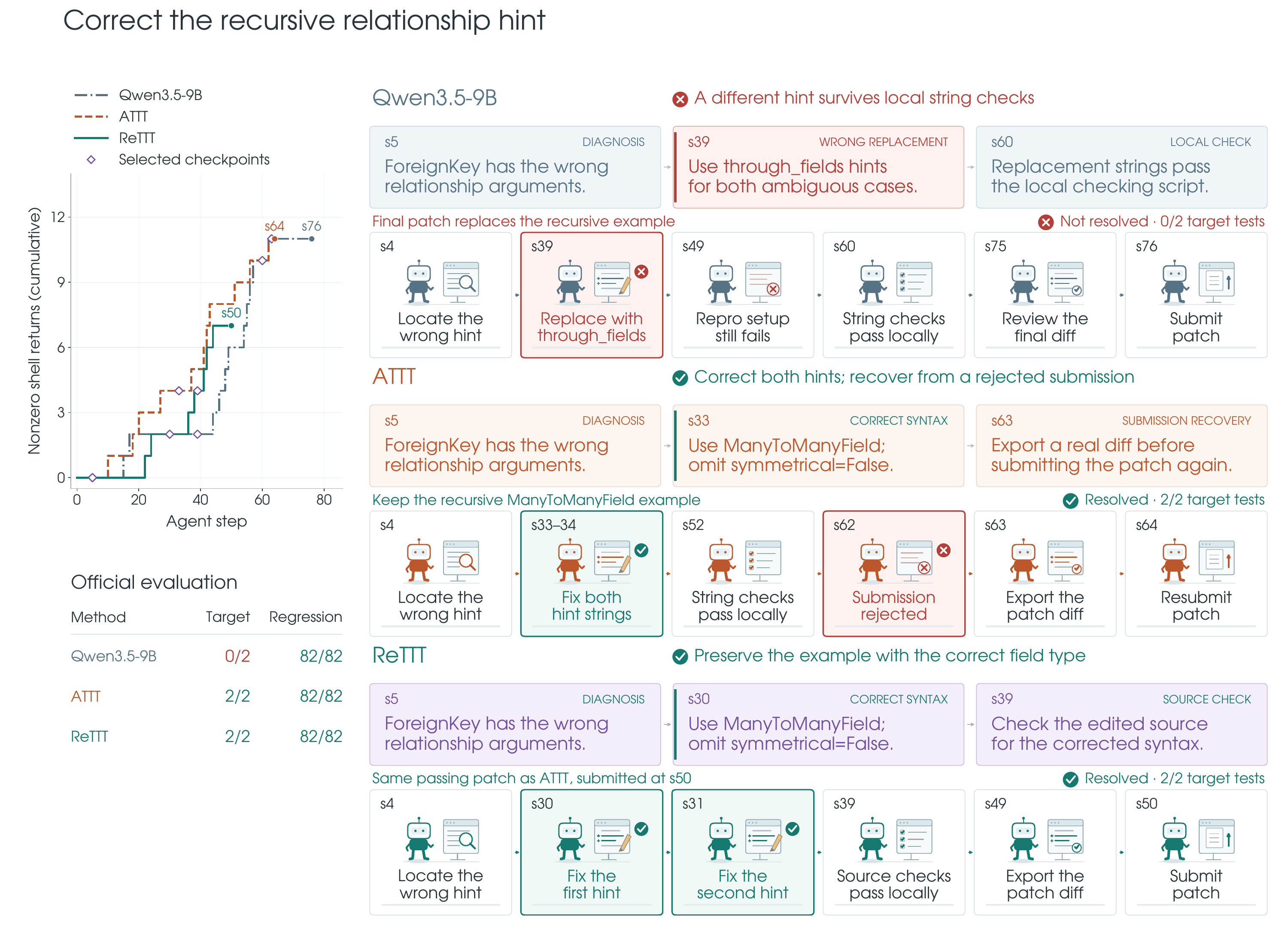}
    \caption{
    \textbf{SWE-bench Lite case: correcting a recursive relationship hint.}
    ReTTT preserves the correct field type and reaches the fully
    passing patch in fewer agent steps than aTTT.
    }
    \label{fig:case_swe_relationship}
\end{figure}

\paragraph{ScienceWorld cases.}
Figures~\ref{fig:case_sci_animal}--\ref{fig:case_sci_friction} illustrate how
procedural adaptation supports multi-step scientific interaction. In the
animal-delivery task, ReTTT identifies the required focus operation and
immediately follows it with pickup and delivery. In the two comparison tasks,
ReTTT gathers evidence from both candidates before committing to a final
choice. In contrast, aTTT either omits a required prerequisite or selects a
candidate before performing the necessary comparison. These cases highlight
the importance of converting intermediate environment feedback into guidance
about what evidence or operation is still required before completing the task.

\begin{figure}[p]
    \centering

    \includegraphics[
        width=0.99\textwidth,
        height=0.38\textheight,
        keepaspectratio
    ]{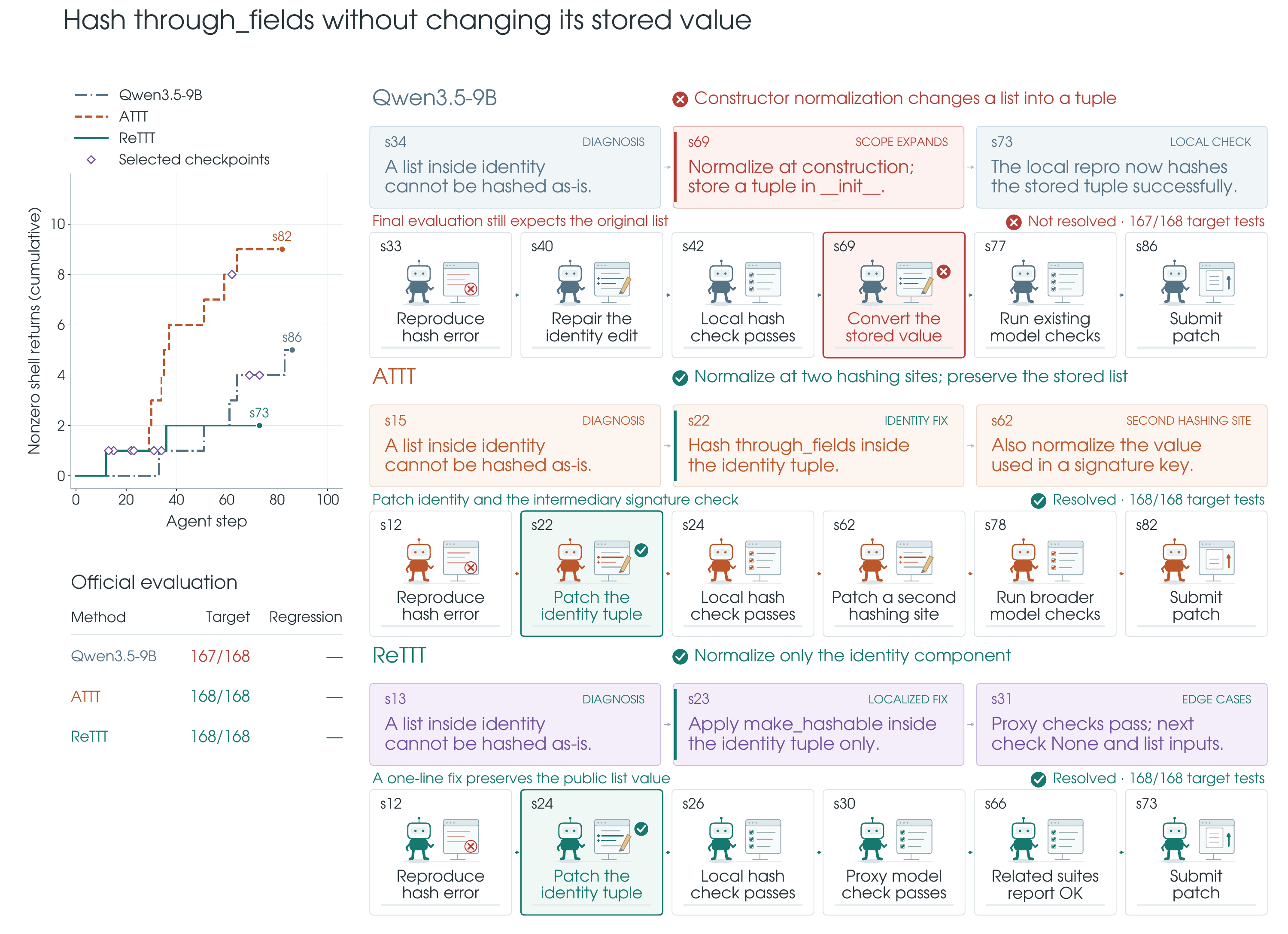}
    \caption{
    \textbf{SWE-bench Lite case: localizing a hashability fix.}
    ReTTT applies the fix only where hashability is required,
    preserving the stored public value and reaching a fully passing solution.
    }
    \label{fig:case_swe_hash}

    \vspace{0.6em}

    \includegraphics[
        width=0.99\textwidth,
        height=0.38\textheight,
        keepaspectratio
    ]{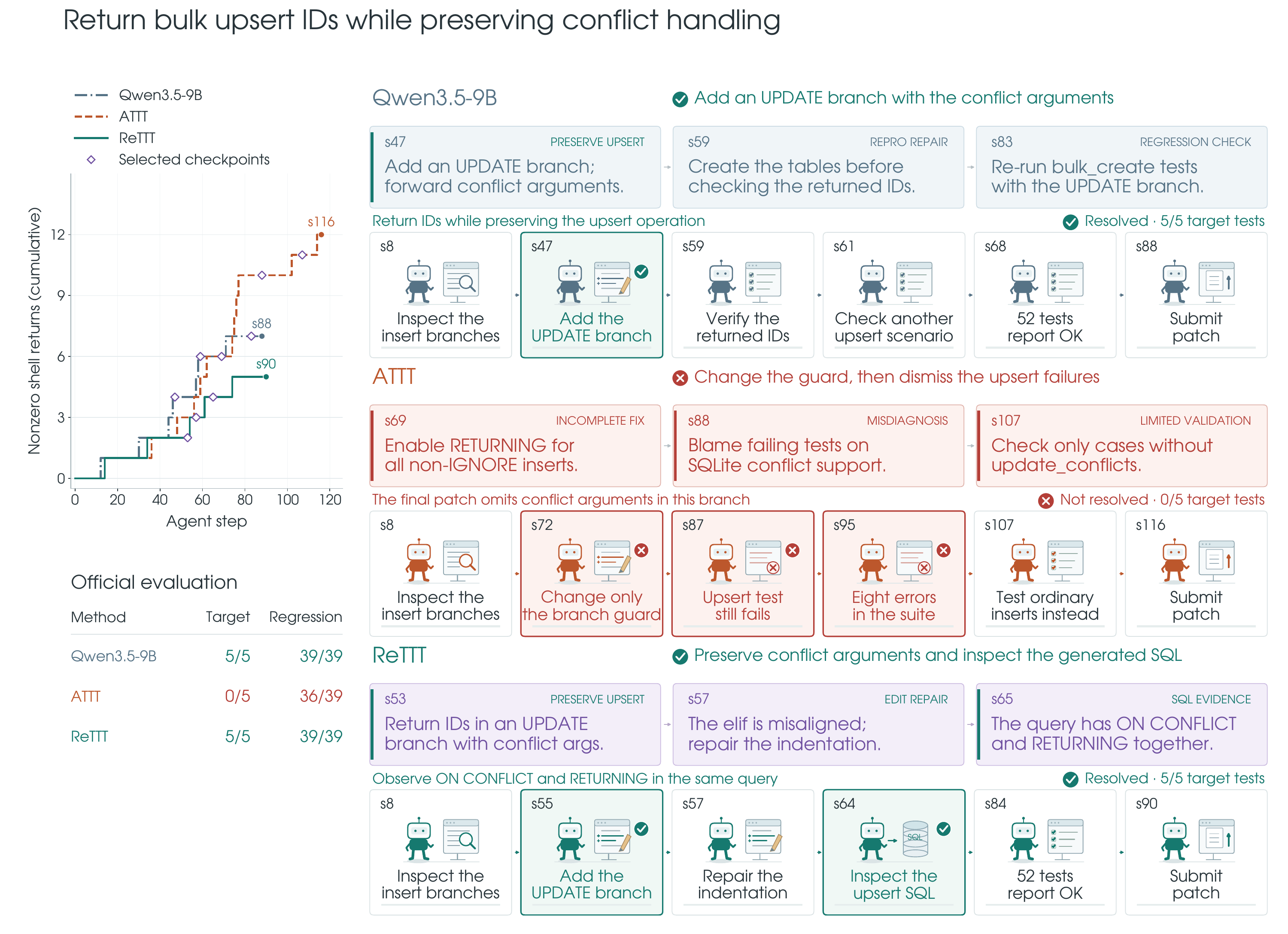}
    \caption{
    \textbf{SWE-bench Lite case: preserving bulk-upsert semantics.}
    ReTTT preserves the conflict arguments and passes all target
    and regression tests, while aTTT converges to an incomplete fix.
    }
    \label{fig:case_swe_upsert}
\end{figure}

\paragraph{SWE-bench Lite cases.}
Figures~\ref{fig:case_swe_relationship}--\ref{fig:case_swe_upsert} show the
same effect in repository-level software engineering. Across all three
examples, ReTTT reaches a fully correct patch with fewer agent
steps than aTTT. Its trajectories make more localized diagnoses and targeted
edits: preserving the correct relationship type, restricting hashability
normalization to the component that requires it, and retaining conflict
arguments when adding the missing bulk-upsert branch. This reduces unnecessary
edits and validation cycles while preserving task constraints. In contrast,
aTTT either requires additional correction steps or, in the bulk-upsert case,
converges to an incomplete fix that introduces regressions.

\end{document}